\documentclass[sigconf]{acmart}
\usepackage{amsthm}

\usepackage{lipsum}
\usepackage{tabularx}
\usepackage{multirow}
\usepackage{algorithmic}
\usepackage[vlined,ruled,linesnumbered]{algorithm2e}
\usepackage{makecell}
\usepackage{float}

\usepackage{graphicx}
\usepackage{subfigure}

\AtBeginDocument{%
  \providecommand\BibTeX{{%
    \normalfont B\kern-0.5em{\scshape i\kern-0.25em b}\kern-0.8em\TeX}}}

\setcopyright{acmcopyright}
\copyrightyear{2024}
\acmYear{2024}
\acmDOI{XXXXXXX.XXXXXXX}

\acmConference[Conference acronym 'XX]{Make sure to enter the correct
  conference title from your rights confirmation emai}{June 03--05,
  2018}{Woodstock, NY}
\acmPrice{15.00}
\acmISBN{978-1-4503-XXXX-X/18/06}

\begin{document}

\title{Class-Balanced and Reinforced Active Learning on Graphs}



\author{Chengcheng Yu}
\affiliation{%
  \institution{Shanghai Polytechnic University}
  \city{Shanghai}
  \country{China}}
\email{ccyu@sspu.edu.cn}

\author{Jiapeng Zhu}
\affiliation{%
  \institution{East China Normal University}
  \city{Shanghai}
  \country{China}
}
\email{jiapengzhu@stu.ecnu.edu.cn}

\author{Xiang Li}
\affiliation{%
 \institution{East China Normal University}
  \city{Shanghai}
  \country{China}
}
\email{xiangli@dase.ecnu.edu.cn}

\renewcommand{\shortauthors}{Trovato and Tobin, et al.}

\begin{abstract}
Graph neural networks (GNNs) have demonstrated significant success in various applications, such as node classification, link prediction, and graph classification. Active learning for GNNs aims to query the valuable samples from the unlabeled data for annotation to maximize the GNNs' performance at a lower cost. However, most existing algorithms for reinforced active learning in GNNs may lead to a highly imbalanced class distribution, especially in highly skewed class scenarios. GNNs trained with class-imbalanced labeled data are susceptible to bias toward majority classes, and the lower performance of minority classes may lead to a decline in overall performance. To tackle this issue, we propose a novel class-balanced and reinforced active learning framework for GNNs, namely, {\itshape GCBR}. It learns an optimal policy to acquire class-balanced and informative nodes for annotation, maximizing the performance of GNNs trained with selected labeled nodes. GCBR designs class-balance-aware states, as well as a reward function that achieves trade-off between model performance and class balance. The reinforcement learning algorithm Advantage Actor-Critic (A2C) is employed to learn an optimal policy stably and efficiently. We further upgrade GCBR to GCBR++ by introducing a punishment mechanism to obtain a more class-balanced labeled set. Extensive experiments on multiple datasets demonstrate the effectiveness of the proposed approaches, achieving superior performance over state-of-the-art baselines.

\end{abstract}



\keywords{Active Learning, Class Balanced, Graph Neural Network, Reinforcement Learning}


\received{20 February 2007}
\received[revised]{12 March 2009}
\received[accepted]{5 June 2009}

\maketitle

\section{Introduction}

Graph neural networks (GNNs) have recently attracted significant attention 
due to their success in various downstream tasks such as node classification \cite{kipf2016semi,hamilton2017inductive}, link prediction \cite{zhang2018link}, and graph classification \cite{morris2019weisfeiler}.
For example, 
the early model GCN \cite{kipf2016semi} aggregates node features in the spectral space using a simplified first-order approximation, while GraphSage \cite{hamilton2017inductive} aggregates features from node neighbors directly in the spatial domain. 
Despite the appealing performance, 
most GNN models typically require a sufficient amount of labeled data for training, which is usually expensive to obtain. 

\begin{figure}[h]
\centering
\begin{minipage}[h]{0.23\textwidth}
\centering
\includegraphics[scale=0.33]{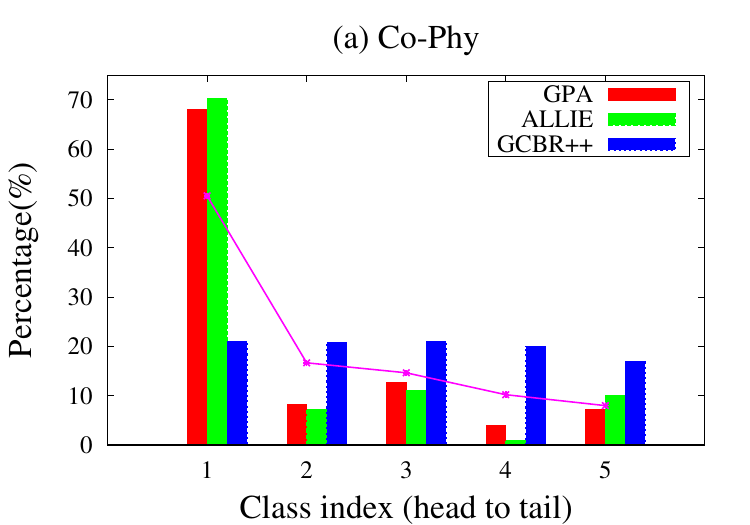}

\end{minipage}
\begin{minipage}[h]{0.23\textwidth}
\centering
\includegraphics[scale=0.33]{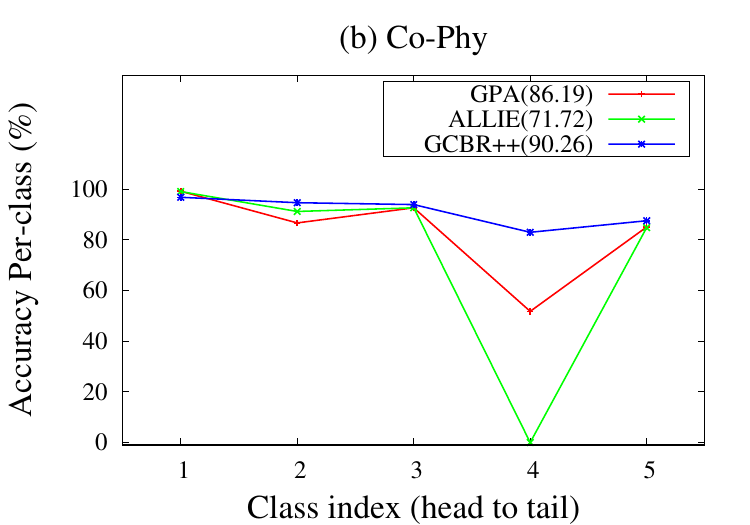}

\end{minipage}

\caption{(a) Class distribution of labeled nodes acquired by GCBR++ is more balanced compared to GPA and ALLIE on Co-Phy dataset. (b) GCBR++ outperforms GPA and ALLIE on overall accuracy. GCBR++ improves the performance of tail classes by a large margin due to the class-balanced and valuable training data.}
\label{fig:fig1}
\end{figure}

Active learning with GNNs emerges as a promising strategy to address this challenge, 
whose goal is to dynamically query the most valuable samples from unlabeled data for annotation to maximize GNNs' performance within a limited budget. 
There have been 
some active learning methods~\cite{cai2017active,gao2018active,chen2019activehne,hu2020graph} for graphs 
proposed and shown effective, 
which introduce different selection criteria to measure the informativeness of each node, such as information entropy, node centrality, and information density. 
However, samples selected by these methods may exhibit a highly imbalanced class distribution, and this imbalance is particularly emphasized in scenarios involving highly skewed classes.
For instance, 
as shown in the pink line of Figure \ref{fig:fig1}(a),
the benchmark dataset \emph{Coauthor\_Physis} (Co-phy) \cite{shchur2018pitfalls},
a co-authorship network of authors categorized into five {classes},
has a highly skewed class distribution that  
is sorted in a decreasing order w.r.t. the number of samples in the corresponding class. 
Specifically, the first class has a significant number of samples (\textit{a.k.a}, head classes), while the remaining classes have only a few samples (\textit{a.k.a}, tail classes). 
We use two SOTA active learning methods GPA~\cite{hu2020graph} and ALLIE~\cite{cui2022allie} (Details will be introduced in Section~\ref{sec:related}), as well as our proposed method GCBR++, to acquire the same number of nodes for annotation on the dataset.
From Figure \ref{fig:fig1}(a),
we observe that the class distributions of labeled nodes acquired by GPA and ALLIE are both highly imbalanced, 
and most nodes are from the head class,
while GCBR++ derives balanced results. 
To further show the effectiveness of the selected nodes, 
we train GCN with these labeled nodes and perform node classification. 
The result is shown in Figure \ref{fig:fig1}(b), where GPA and ALLIE obtain poorer results, especially on the tail class 4 with the least number of labeled nodes. In contrast, our method achieves robust performance across different classes.

The class-imbalanced labeled training data can lead to the class-imbalanced problem in the classification results~\cite{chawla2002smote,cui2019class}. 
Since typical GNNs are designed without considering the problem,
training GNNs with class-imbalanced labeled data could introduce a prediction bias toward majority classes, resulting in overall performance degradation.
Meanwhile, class imbalance is prevalent in real-world applications, 
such as fraud detection \cite{liu2021pick}, citation networks \cite{wang2020network}, and social networks \cite{shi2023over}.
Therefore, it is crucial for active learning to select class-balanced and valuable samples to annotate and avoid class-imbalanced problems. 
Although
active learning for imbalanced classes has been well studied in computer vision \cite{aggarwal2020active,kothawade2021similar,bengar2022class,zhang2022galaxy,zhang2023algorithm} recently,
most existing methods are based on the i.i.d assumption, and directly applying them to graphs might be unsuitable or ineffective. 
This further prompts our investigation on class-balanced active learning for GNNs.


In this paper, we propose a class-balanced active learning approach for graphs, namely, GCBR. It adopts the RL framework and learns an optimal policy 
to query class-balanced and informative samples to annotate, 
maximizing the performance of GNNs trained with the selected labeled nodes.
Specifically, 
our method formalizes class-balanced active learning as a Markov Decision Process (MDP) and learns the optimal query strategy. 
The state is defined based on the current graph status characterized from both node informativeness and class balance; 
the action is to select a node to label at each query step;
the reward function is designed with the performance gain of GNNs trained with the selected nodes and also a pre-defined class diversity score, which can improve both performance and class balance. 
To obtain a more class-balanced labeled set, 
we further upgrade GCBR to GCBR++ 
by incorporating a punishment mechanism, 
which adds a penalty term in the reward function and
enforces nodes in the minority classes to be selected.
For more stable and effective training, Advantage Actor-Critic (A2C) algorithm \cite{mnih2016asynchronous} is used to learn the query policy,
where
the actor and critic networks are composed of two GCNs, considering both informativeness and the inter-dependent connections between nodes. 
Finally, 
our main contributions in the paper are summarized as follows:
\begin{itemize}

\item We propose {GCBR}, a novel class-balanced active learning approach for GNNs. To the best of our knowledge, we are the first to introduce class balance to reinforced active learning on graphs.
\item We design an effective reward function that can strike the trade-off between classification performance and class balance.
We also introduce 
class-balance-aware state space for sampling informative nodes. 

\item We conduct extensive experiments on six benchmark datasets to show that our methods can obtain a more class-balanced labeled set,
which leads to 
comparable or better classification results than SOTA competitors. 

\end{itemize}

\section{Related Work}
\label{sec:related}
\subsection{Active Learning on Graphs}
Active learning has been widely studied in various domains, such as computer vision and natural language processing. 
Several works focus on active learning for graph-structured data recently \cite{cai2017active,gao2018active,chen2019activehne,wu2019active,zhang2021alg}. 
For example, AGE \cite{cai2017active} measures node informativeness by a linear combination of three criteria, including information entropy, density, and centrality. It finds the most informative nodes from unlabeled nodes.
ANRMAB \cite{gao2018active} extends AGE, employs the same criteria, and dynamically adjusts linear combination weights by a multi-armed bandit mechanism.
ActiveHNE \cite{chen2019activehne} employs the multi-armed bandit mechanism to tackle active learning on heterogeneous graphs.
However, these methods 
overlook the connections between nodes in the graph-structured data. 
Several studies on active learning leverage reinforcement learning to acquire a labeling policy and parameterize the policy network as GNNs, which could model node interactions. 
GPA \cite{hu2020graph} trains a GNN-based policy network with reinforcement learning to select nodes one by one.
ALLIE \cite{cui2022allie} extends GPA, an active learning method for large-scale class-imbalanced scenarios. 
BIGENE \cite{zhang2022batch} proposes a multi-agent reinforced active learning framework to query multiple samples at each time.
However above active learning methods may exhibit a highly imbalanced
class distribution, as shown in Figure 1. GNNs trained with class-imbalanced labeled data are prone to be biased toward major classes compared to minority classes, resulting in overall performance degradation. It's significant to get class-balanced and informative labeled nodes by active learning for training the unbiased GNNs classification with selected nodes.

\subsection{Class-Imbalanced Problem}

The class-imbalanced problem has been extensively studied \cite{mullick2019generative,chawla2002smote,cui2019class}. 
The goal is to train an unbiased classifier from a class-imbalanced distribution, where the majority classes have many more training instances than the minority classes.
Some pioneering works have recently explored class-imbalanced learning for graphs \cite{ma2023class}, including data-level and algorithm-level approaches. Data-level approaches modify the training data to achieve a more class-balanced learning environment. For example, GraphSMOTE \cite{zhao2021graphsmote} and GraphENS \cite{park2022graphens} generate synthetic minority samples and edges connected with them as data augmentation for minority classes. GraphSR \cite{zhou2023graphsr} leverages the large number of unlabeled nodes in graphs as data augmentation directly. 
Algorithm-level approaches modify the learning algorithms of the classifier to tackle the class-imbalanced issue. ReNode \cite{chen2021topology} and TAM \cite{song2022tam} modify loss functions by raising the weights of minority classes to address class imbalance. 
This paper aims to prevent the class imbalance issue that could arise during active learning. Class balance, as one of the active learning objectives, is essential in imbalanced datasets.

\subsection{Class-Imbalanced Active Learning}
In recent years, active learning for class-imbalanced has received increasing attention in computer vision \cite{aggarwal2020active,kothawade2021similar,bengar2022class,zhang2022galaxy,zhang2023algorithm}. 
For example, CBAL \cite{bengar2022class} proposes an optimization-based method that aims to balance classes for image classification, which can be combined with other criteria.
SIMILAR \cite{kothawade2021similar} select samples with gradient embeddings for annotation, which are most similar to previously collected minority samples while most dissimilar to out-of-distribution ones.
TAILOR \cite{zhang2023algorithm} proposes a selection strategy for deep active learning, which uses novel reward functions to gather class-balanced samples.
Nevertheless, directly applying these methods to graphs might be unsuitable or ineffective due to their i.i.d assumption. In contrast, we focus on graph-structured data with the goal of class-balance active learning.

\section{Problem Definition}
Let $G = (V, E)$ denotes a graph, where $V$ is the node set, and $E$ is the edge set. $A\in \mathbb{R}^{N \times N}$ is the adjacency matrix, and $X\in \mathbb{R}^{N \times d}$ is the node feature matrix.
Each node $v\in V$ has a label $c(v)\in \{1,...,m\}$, $m$ is the number of node classes.
The node set is divided into three subsets, including $V_{train}$, $V_{valid}$ and $V_{test}$. In traditional semi-supervised node classification, the labels of a subset $L \subseteq V_{train} $ are given. The task is to learn a classification $f$ with the graph $G$ and labels $L$ to predict the node labels in $V_{test}$. We denote the class distribution of $L$ as $\{C_{1},...,C_{m}\}$, where $C_i =|\{v|v \in L ,c(v) =i\}|$ is the node number of the $i$-th class in $L$. Besides, we use an imbalance ratio $\rho$ to measure the balance degree of the class distribution. 
\begin{equation}
  \rho = \frac{min\{C_{1},...,C_{m}\}}{max\{C_{1},...,C_{m}\}} 
\label{equation:imbRatio}
\end{equation}

For class-balanced active learning on graphs, the labeled subset is initialized as an empty set $L^0=\phi$. A query budget $B$ is given, and we sequentially acquire the labels of $B$ samples. At each step $t$, we select an unlabeled node $v^t$ from $V_{train}\backslash L^{t-1}$, query the oracle for its labels, and expand labeled dataset $L^t=L^{t-1} \cup \{v^t\}$, then the GNN classification $f(G,L^t)$ is trained with $L^t$ for one more epoch. This process is repeated until the $B$ is used up. Finally, we continue training the classification $f(G,L^B)$ until convergence. Our goal is to gather class-balanced labeled subset $L^B$ ($\rho$ should be as close to 1 as possible) and maximize the performance of the GNN classification $f(G,L^B)$ on $V_{test}$.

\section{Methodology}

\begin{figure*}[h]
  \centering
  \includegraphics[scale=0.45]{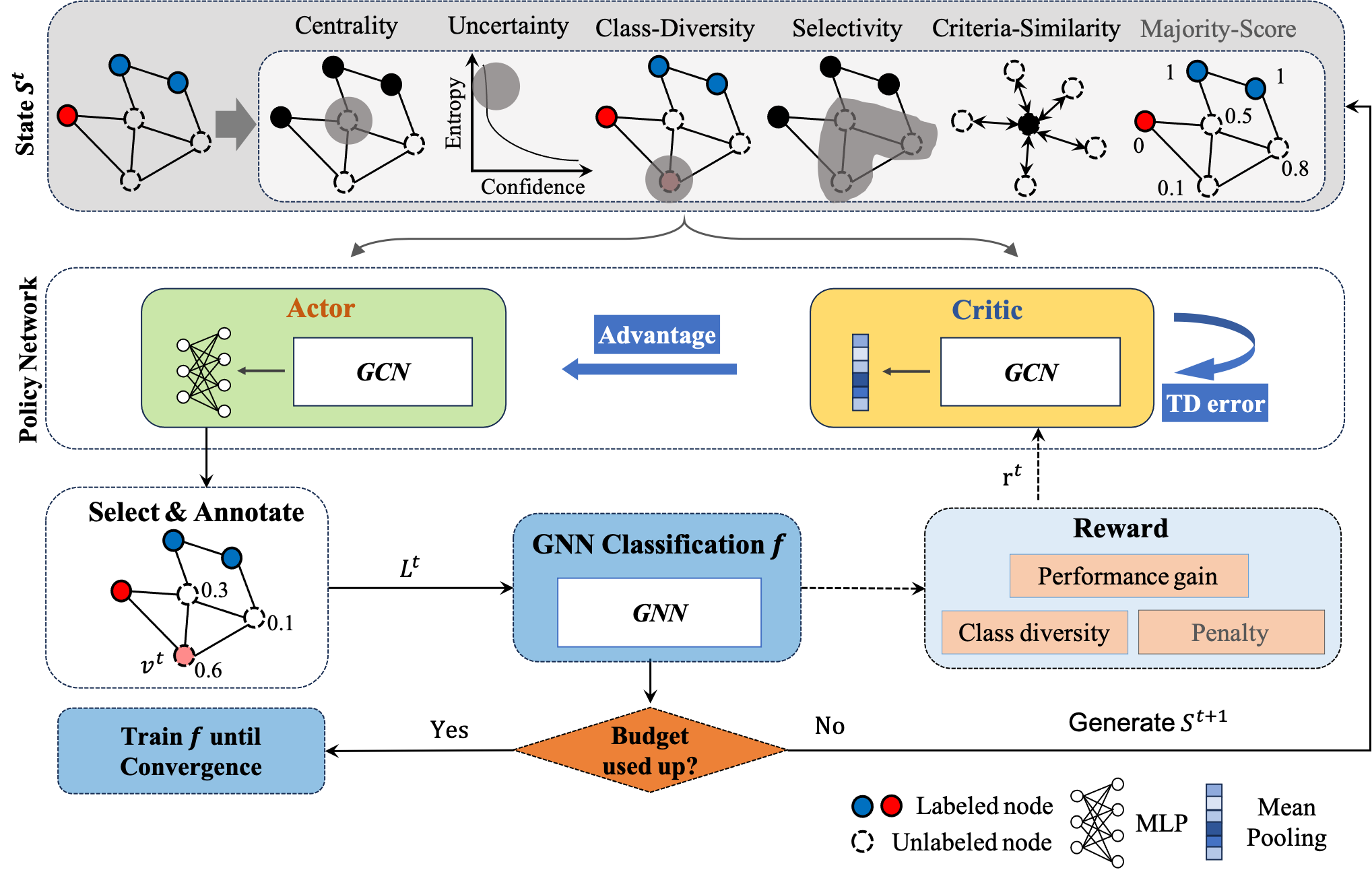}
  \caption{The framework of GCBR and GCBR++. Blue and red nodes denote the classes of labeled nodes, and blank nodes are unlabeled nodes. The policy will query a more class-balanced and valuable node for annotation at each step.}
  \label{framework}
\end{figure*}

In this section, we describe our proposed models. 
We formalize the problem of class-balanced active learning on graphs as a Markov Decision Process (MDP).
Specifically,
the state is defined based on the current condition of a graph $G$ and the trained GNN classification $f$. 
We denote the state at step $t$ as $S^t$. 
The active learning policy $\pi_{\theta}$ parameterized by $\theta$ takes an action by selecting the next node to query.
To improve the performance of the classification and also the class balance of labeled nodes, the instant reward $r^t$ at step $t$ is designed based on the performance gain of the GNN and the class diversity. 
We show an overview of the proposed framework of GCBR in Figure \ref{framework}. 
Initially, the labeled set $L^0$ is set to be empty.
At step $t$, we first use the output of the GNN classification $f(G,L^{t-1})$ and $G$ to update the state $S^t$. A node $v^t$ is sampled from $V_{train} \backslash L^{t-1}$ in terms of the policy $\pi_{\theta} (\cdot | S^t)$ for annotation, and is added to the label set $L^t = L^{t-1} \cup \{v^t\}$. Then the GNN classification $f(G,L^{t})$ is trained for one epoch, then evaluate $f(G,L^{t})$ on $V_{valid}$, which is used to generate the graph state $S^{t+1}$ for next step. When the budget $B$ is used up, we stop querying and train classification $f(G,L^{B})$ until convergence. 
The A2C algorithm is employed to train the policy network, and a batch training mode is designed to update the parameters of the policy network at a frequency of every $F$ actions. To acquire more class-balanced labeled nodes, we upgrade GCBR to GCBR++ by incorporating a punishment mechanism. 
In the following, we discuss the major components in detail.

\subsection{State}
\label{states}
The state of graph $G$ at step $t$ is denoted as a matrix $S^t$, where each row $s_{v}^t$ represents the state representation of node $v$. 
To obtain class-balanced and valuable labeled nodes, the state is characterized from the following five factors/dimensions: 

\textbf{Centrality:} 
Nodes in graphs are not i.i.d. but connected with links. 
The centrality of a node reflects the extent to which the node influences others. 
Various node centrality metrics have been proposed, such as degree centrality and betweenness centrality \cite{brandes2001faster}. 
It has been shown that using PageRank as the centrality criterion for selecting nodes in active learning outperforms others \cite{cai2017active,zhang2022batch}.
Hence, we adopt PageRank to measure the centrality of node $v$:
\begin{equation}
  s_{v}^t( 1 ) = \delta \sum_{u} A_{vu} \frac{s_{u}^t(1)}{\sum_{k}A_{uk}} + \frac{1-\delta}{N},
\label{equation:s1}
\end{equation}
where 0 < $\delta$ < 1 is the damping parameter.

\textbf{Uncertainty:} The uncertainty-based selection strategy finds extensive application in studies on active learning. The less confident a node's predicted label is, the more likely the classification is to make mistake on the node, the more beneficial the node is to be labeled. For node $v$,
we employ the entropy of its prediction as the measure of uncertainty:
\begin{equation}
  s_{v}^t( 2 ) = - \frac{\sum_{i=1}^{m} \hat{y}_{v}^{t}(i) log( \hat{y}_{v}^{t}(i) )}{log( m )}, 
\label{equation:s2}
\end{equation}
where $\hat{y}_{v}^{t}(i)$ is the probability of node $v$ belonging to the $i$-th class predicted by the GNN classification $f$ at step $t$. We divide the entropy by $log(m)$ for normalization.

{\textbf{Class-Diversity:}} Inspired by~\cite{zhang2023algorithm}, we propose class diversity to measure a node's importance in maintaining class balance. Intuitively,
the more nodes in a class has been selected,
the less nodes in the class are to be selected.
Therefore,
in each step,
we assign an inverse weight to each class based on the number of samples that are already selected. Formally, the class diversity of node $v$ is computed by the weighted sum of class probabilities: 
\begin{equation}
s_{v}^t( 3 ) = \sum_{i=1}^{m} \frac{\tilde{y}_{v}^t(i)}{max(1, C_i^{t-1})}, 
\label{equation:s3}
\end{equation}
where $\tilde{y}_{v}^t(i)$ denotes the probability of node $v$ in the $i$-th class. If node $v$ has been labeled, $\tilde{y}_{v}^t(i) = y_{v}(i)$, and $y_{v}(i)$ refers to the $i$-th element of the ground-truth one-hot label vector $y_{v}$; otherwise, $\tilde{y}_{v}^t(i) = \hat{y}_{v}^{t-1}(i)$. 
Further, $C_i^{t-1}$ is the number of nodes with the $i$-th label in $L^{t-1}$. To avoid the invalidity of inverse weight for classes with zero nodes selected,
we set the inverse weight to be $\frac{1}{max(1, C_i^{t-1})}$.

\textbf{Selectivity:} 
Following \cite{hu2020graph}, we use an indicator variable to represent whether a node has been selected/labeled. This is because we expect the policy network to focus more on unlabeled samples. 
Specifically, for node $v$, we define 
\begin{equation}
    s_{v}^t( 4 ) = \vmathbb{1} \{ v \in L^{t-1} \}
\label{equation:s4}
\end{equation}
where $s_{v}^t(4) = 1$ if node $v$ has been labeled; otherwise, $s_{v}^t(4) = 0$.

{\textbf{Criteria-Similarity:}} The above four factors are prone to select unlabeled central and uncertain nodes that can contribute to class diversity.
We expect that the next selection will also derive similar node.
Therefore,
we propose a novel metric to measure the criteria similarity between unlabeled and labeled nodes. 
We concatenate the four factors mentioned above as the criteria representation of each node. Then the Criteria-Similarity of node $v$ is calculated as: 
\begin{equation}
  s_{v}^t( 5 ) = \underset{u\in L^{t-1}}{\min} d( \hat{S}_v^t, \hat{S}_u^t ),
\label{equation:s5}
\end{equation}
where $\hat{S}_v^t$ is the criteria representation of node $v$ at step $t$ and $d(\cdot,\cdot)$ denotes the Euclidean distance.
Note that previously, active learning aims to select nodes that are most ``dissimilar'' to labeled nodes to label, where the dissimilarity is measured from the perspective of vanilla node features. However, in our case, we use
criteria similarity to enforce the selection of nodes similar to the training set w.r.t. the selection criteria, i.e.,  
unlabeled central and uncertain nodes that can contribute most to class diversity.

Finally, we concatenate the above five metrics to form the state representation $s_v^t$ for each node $v$ at step $t$. Then, the graph state matrix $S^t$ will be passed into the policy network to generate the action probabilities.

\subsection{Action}
At step $t$, the action is generated from the policy network $\pi_{\theta} (\cdot | S^t)$ based on the current state $S^t$, which is used to to select a node $v^t$ from $V_{train}\backslash L^{t-1}$. In this work, the policy network represents the probability distribution of the action space, which includes all the unlabeled nodes in the graph. (See Section \ref{Policy Network Architecture} for policy network.)

\subsection{Reward}
Our goal is not only to improve the performance of GNNs but to obtain class-balanced labeled nodes. 
Our reward is thus designed based on the performance gain on the validation set and the class diversity. 
At the $t$-th step, 
suppose that node $v^t$ is selected by the policy network for annotation, and the reward function is given by:
\begin{equation}
 r^{t} = \alpha \cdot g(v^t)+ (1-\alpha) \cdot h(v^t), \label{equation:reward} 
\end{equation}
where the scaling factor $\alpha\in [0,1]$ controls the importance of the two terms.
Here, $g(v^t) = \mathcal{M}( f(G,L^t), V_{valid})-\mathcal{M}( f(G,L^{t-1}), V_{valid}) $ represents the performance gain on the validation set after node $v^t$ is added to the training set. $\mathcal{M}$ is the performance evaluation function \emph{Macro-F1}, which is the average of the F1 score per class. $h(v^t) = \sum_{i=1}^{m} \frac{y_{v^t}(i)}{max(1,C_i^{t-1})}$ indicates the class diversity score, and $y_{v^t}$ is the one-hot label vector of node $v^t$.
In particular, 
when the class of $v^t$ has less nodes selected,
$h(v^t)$ will lead to a larger reward.


\subsection{Policy Network Architecture}
\label{Policy Network Architecture}
Advantage Actor-Critic (A2C) is a deep reinforcement learning algorithm that combines policy-based and value-based methods. 
The policy network architecture consists of the actor and the critic networks.
The actor is used to select a node for annotation, while the critic provides value estimation of a given state. To fully utilize the graph structural information, we parameterize the actor and critic as two $L$-layer GCNs. 
Specifically, the propagation rule for layer $l$ in the GCN of the actor is:
\begin{equation}
H_{actor}^{(l+1)} = \sigma ( \tilde{D} ^{-\frac{1}{2}} \tilde{A} \tilde{D} ^{-\frac{1}{2}} H_{actor}^{(l)} W_{actor}^{(l)} )
\end{equation}
where $\tilde{A} = A+I$ is the adjacency matrix with self loops and $\tilde{D}$ is the degree matrix of $\tilde{A}$. $W_{actor}^{(l)}$ is the weight matrix and $\sigma$ is the activation function \texttt{ReLU}. Here, $H_{actor}^{(0)}=S^t$ is the initial input feature. The GCN of the critic has the similar propagation rule $H_{critic}^{(l+1)}$, whose details are omitted due to the page limitation.

In the actor part, 
after $L$ layers,
the final output $H_{actor}^{(L)}$ will be further fed into a linear layer to get a score for each node. Then, the probability distribution $\pi_{\theta}$ is computed by normalizing these scores using the \texttt{softmax} function. 
\begin{equation}
\pi_{\theta} (\cdot | S^t )= \texttt{softmax} ( W H_{actor}^{(L)} +b).
\end{equation}
An action $a$ (i.e., node $v^t$) which satisfies $V_{train}/ L^{t-1}$ is sampled from the distribution $\pi_{\theta}$ to encourage exploration during training, while we greedily select the action corresponding to the largest probability during inference. The selected node $v^t$ is then annotated.

In the critic part, we apply a mean pooling on $H_{critic}^{(L)}$, 
which derives the state value function $V_{\phi} ( S^t )$: 
\begin{equation}
V_{\phi} ( S^t )= \texttt{MeanPooling}(H_{critic}^{(L)}).
\end{equation}

\subsection{Policy Training}
By using the advantage function as a baseline to reduce the variance of the policy gradient estimation, A2C can update the policy more stably. To further improve training efficiency, we design a batch training mode to update the weights of the critic and actor networks at a frequency of every $F$ actions. 

At step $t$, the actor chooses an action $v^t$ based on the current state $S^t$, and the node $v^t$ is added into $L^{t-1}$ after labeling. The classification $f(G,L^t)$ is trained for one epoch, and the instant reward $r^{t}$ is then obtained. After that, we can generate the next state $S^{t+1}$.

The critic predicts state value $V_{\phi} (S^{t})$ for the given state $S^{t}$. The target value $V^{*}(S^{t})$ is obtained via the Bellman equation:
\begin{equation}
V^{*}(S^{t}) =r^t + \gamma V_{\phi} (S^{t+1})
\label{equation:target}
\end{equation}
where $\gamma$ is the discount factor for balancing instant rewards and future rewards. 
The critic is updated by minimizing the Temporal Difference (TD) error:
\begin{equation}
J(\phi) =\frac{1}{F} \sum_{t-F}^{t} (V^{*}(S^{t}) -V_{\phi} (S^{t}))^2.
\label{equation:critic}
\end{equation}

The advantage function calculates how better taking that action at a state is compared to the average value of the state. TD error is employed as an unbiased estimate of the advantage function $A$. Formally, we denote 
\begin{equation}
A(S^t,v^t)= V^{*}(S^{t})-V_{\phi} (S^{t}).
\label{equation:advantage}
\end{equation}

The objective of the actor is to maximize the expected cumulative reward. We update the actor network parameter $\theta$ using policy gradient ascent based on the advantage function. The loss function of the actor is given as:
\begin{equation}
J(\theta) = - \frac{1}{F} \sum_{t-F}^{t} log\pi_\theta(v^t|S^t)\cdot A(S^t,v^t).
\label{equation:actor}
\end{equation}

We next summarize the key steps of training the policy network as follows.
For each step $t$, the actor selects an action $v^t$ based on the current state. Then, the classification model GNN will be trained with cross-entropy loss for one epoch. An instant reward is calculated using Equation \ref{equation:reward}. Target value and advantage function are computed based on the current and the next state value estimated by the critic according to Equation \ref{equation:target} and \ref{equation:advantage}, respectively. The actor and critic networks are updated with a batch of $F$ transitions using the advantage function by optimizing Equation \ref{equation:critic} and \ref{equation:actor}, respectively. $B$ and $F$ are typically set as integer multiples of the number of classes in the training graph, and $B=kF, k\in \{1,2...\}$. Due to the space limit, we give the detailed pseudo-code for policy training in Appendix A.

We adopt a transferable active learning scheme to train the policy network similar to GPA \cite{hu2020graph}. To clarify, we use a signal graph $G_s$ with full label information to train the policy, which is used to test the other target graph $G_t$ without adaptation.

\subsection{GCBR++}

To obtain a more class-balanced labeled set, we further upgrade GCBR to GCBR++ by incorporating a punishment mechanism. This mechanism includes a penalty term in the reward function and introduces a majority score in the state space. 
The motivation is as follows. 
If a majority node is selected for annotation, 
the reward will be reduced by a penalty score.
Then we modify the reward function in Equation \ref{equation:reward} into:
\begin{align}
r^{t} &= \alpha \cdot g(v^t)+ (1-\alpha) \cdot h(v^t) - \eta  \cdot  \vmathbb{1} \{ c(v^t)\in C_{major}^{t-1} \}  \label{equation:reward+} 
\end{align}
where $\eta > 0$ is a penalty score, which is a hyperparameter. $\vmathbb{1} \{ c(v^t)\in C_{major} \}$ is an indicator variable to represent whether the class of $v$ belongs to the majority class. $C_{major}^{t-1}= \{i \, | \, C_{i}^{t-1} \ge B/m\}$ denotes the set of the classes belonging to the majority class at step t-1.

Meanwhile, the metric of \emph{Majority-Score} will be added to the state of the graph as the sixth characterization factor, which can help the policy identify nodes that are in minority classes.

\textbf{Majority-Score:} 
If an unlabeled node is more likely to belong to majority classes, it is expected to be not selected. We calculate the probability of node $v$ belonging to the majority classes as:  
\begin{equation}
s_{v}^t( 6 ) = \sum_{i \in C_{major}^{t-1}}^{} \tilde{y}_{v}^t(i),
\label{equation:s6}
\end{equation}
where $\tilde{y}_{v}(i)$ indicates the predicted probability score of node $v$ in the $i$-th class.

\section{Experiments}

\subsection{Experiment Setup}
\subsubsection{Datasets}
We use six widely used benchmark datasets, including Cora, Citeseer, Pubmed, Reddit, Coauthor-CS (Co-CS), and Coauthor-Physics (Co-Phy) \cite{shchur2018pitfalls}. 
The first three datasets are citation graphs with nodes as documents and edges as citations. 
Reddit is an online forum dataset where nodes represent posts, and two posts are connected with an edge if at least two users comment or post them. 
Co-CS and Co-Phy are co-authorship networks that connect author nodes by co-authored works. 
The statistics of the datasets are shown in Table \ref{tab:datasets}.
In our experiments, we follow \cite{hu2020graph} to process the Reddit dataset. We train GCBR, GCBR++, GPA, and ALLIE on the Cora dataset and evaluate them on the other five datasets together with other baselines.

\begin{table}
\setlength{\tabcolsep}{3pt}
    \caption{The statistics of the datasets.}
    \centering
    \begin{tabular}{ccccccc}
    \hline 
         Dataset & Nodes & Edges & Features & Classes  & Imb-ratio \\
         \hline 
        Cora &  2708& 5278 &1433& 7  & 0.22 \\
        Citeseer & 3327 & 4676 & 3703 & 6   &0.38\\
        Pubmed & 19718 & 44327 &500  & 3  &0.52\\
        Reddit & 4584 & 19460 & 300 & 10  &0.67\\
        Co-CS & 18333 & 81894 & 6805 & 15  &0.03\\
        Co-Phy & 34493 & 247962 & 8415 & 5  &0.15\\
       \hline 
    \end{tabular}
    \label{tab:datasets}
\end{table}

\subsubsection{Baselines}
We compare our proposed methods with the following baselines. We implement GCBR and GCBR++ with PyTorch and all experiments were conducted on a server with NVIDIA Tesla V100 PCIE 40GB GPU.

\textbf{(1) Random:} In each step, randomly select a node for annotation.

\textbf{(2) AGE\footnote{https://github.com/vwz/AGE.} \cite{cai2017active}:} AGE selects samples with the highest weighted sum of three criteria, including entropy, density and centrality. 

\textbf{(3) ANRMAB\footnote{We utilize the implementation of ANRMAB from GPA directly.} \cite{gao2018active}:} ANRMAB utilizes the same criteria as in AGE and adjusts the weights of linear combination dynamically by a multi-armed bandit mechanism.

\textbf{(4) GPA\footnote{https://github.com/ShengdingHu/GraphPolicyNetworkActiveLearning} \cite{hu2020graph}:} GPA formulates active learning on graphs as an MDP and trains a GNN-based policy network to learn the optimal query strategy with reinforcement learning. 

\textbf{(5) ALLIE\footnote{As the code of ALLIE is not provided, we implement the active learning part based on their paper.} \cite{cui2022allie}:} ALLIE extends GPA, utilizes a RL agent with imbalance-aware reward function to sample nodes from each class in the active learning part, 
and employs the algorithm-level technology and the graph coarsening strategy to address the class-imbalance problem for large-scale graphs.

We also notice that a recent work
BATCH \cite{zhang2022batch} is similar to our topic.
However, the code of the method is not publicly released. 
Further, the results in the paper are visualized but not reported in numerical values, so we cannot replicate the results.
For fairness, we do not take it as our baseline.

\subsubsection{Evaluation Metrics.}
We use Micro-F1, Macro-F1, and imbalance ratio (imb-ratio) as the evaluation metrics. The Micro-F1 and the Macro-F1 are the common metrics in GNN literature \cite{yang2016revisiting}. 
The imbalance ratio is usually in class-imbalance problem literature \cite{zhou2023graphsr}, calculated by Equation \ref{equation:imbRatio}. The more balanced the labeled dataset is, the higher the imbalance ratio.
We use 1000 nodes as the test set and randomly sample 500 nodes from the remaining nodes as the validation set. We run 50 independent experiments with different classification network initializations and report the average classification performance on the test set and the average imbalance ratio.

\subsubsection{Hyperparameter Settings.}
\label{Parameter Settings}
The actor and the critic networks are implemented as two two-layer GCNs with a hidden layer of 8, and they all use Adam \cite{kingma2014adam} as the optimizer with a learning rate of 0.001. 
The policy network is trained for a maximum of 4000 episodes with a batch size of 5, an update frequency of 7, and a training budget of 35. We set the hyperparameter $\alpha = 0.5$. And we set the hyperparameter $\eta = 0.05$ in GCBR++.  
We implement the classification network as a two-layer GCN with a hidden layer size of 64. We train it with Adam optimizer with a learning rate of 0.01. To evaluate the active learning method, we set the test budget on each graph as (20 $\times$ \#classes), which is the common setting in the GNN literature \cite{kipf2016semi}, and the active learning literature \cite{cai2017active,cui2022allie}.

\subsection{Experiment Results and Analysis}

\subsubsection{Node Classification Performance}

\begin{table*}
\caption{Node classification performance (Macro-F1, Micro-F1) and Imb-ratio of labeled nodes. The best and 2nd best are noted in bold font and underlined, respectively.}
\renewcommand{\arraystretch}{1.1}
\setlength{\tabcolsep}{2pt}
    \centering
    \begin{tabular}{ccccccccccccccccc}
    \hline 
        \multirow{2}{*}{Method} & \multicolumn{3}{c }{Citeseer} & \multicolumn{3}{c }{Pubmed}& \multicolumn{3}{c }{Reddit}& \multicolumn{3}{c }{Co-CS}& \multicolumn{3}{c }{Co-Phy}\\
         & {\footnotesize Micro-F1} & {\footnotesize Macro-F1}  & {\footnotesize Imb-ratio} & {\footnotesize Micro-F1} & {\footnotesize Macro-F1}  & {\footnotesize Imb-ratio} & {\footnotesize Micro-F1} & {\footnotesize Macro-F1}  & {\footnotesize Imb-ratio} & {\footnotesize Micro-F1} & {\footnotesize Macro-F1}  & {\footnotesize Imb-ratio} & {\footnotesize Micro-F1} & {\footnotesize Macro-F1}  & {\footnotesize Imb-ratio}  \\
         \hline 
        Random&70.96&64.39&0.27&77.55&76.78&0.50&92.35&92.17&0.47&	90.27&77.35&0.03&90.76&82.82&0.14\\
        AGE&72.21&\textbf{67.72}&0.46&80.13&79.23&0.61&92.66&92.58&0.26&92.87&91.62&0.14&92.82&88.11& 0.50\\
        ANRMAB &70.64&64.83&0.34&77.14&75.88&0.47&91.10&91.02&	0.39&	92.55&	89.92&	0.08&	90.79&	82.95&	0.14\\


        GPA&72.16&	67.19&	0.34&	80.13&	78.89 & 	0.36&	92.81&	92.70&	0.47&	92.93&	81.74&	0.02&	91.99&	86.18&	0.06\\

        ALLIE&\textbf{72.74}&	67.19&	0.28&	79.20&	78.29&	0.45&	92.73&	92.62&	0.35&	93.28&	83.42&	0.02&	89.87&	71.72&	0.01\\

         \hline 
        GCBR&\underline{72.50}&	\underline{67.51}&	\underline{0.45}&	\textbf{81.76}&	\textbf{80.46}&	\underline{0.66}&	\textbf{93.64}&	\textbf{93.51}&	\underline{0.47}&	\textbf{94.32}&	\textbf{93.69}&	\underline{0.52}&	\textbf{94.59}&	\textbf{90.47}&	\underline{0.61}\\
        GCBR++&72.22&	67.11&	\textbf{0.57}&	\underline{81.24}&	\underline{80.11}&	\textbf{0.82}&	\underline{93.49}&	\underline{93.33}&	\textbf{0.66}&	\underline{93.57}&	\underline{93.19}&	\textbf{0.69}&	\underline{94.36}&	\underline{90.26}&	\textbf{0.81}\\

       \hline 
    \end{tabular}
    
    \label{tab:performance}
    
\end{table*}

Table \ref{tab:performance} summarizes the performance results of all methods on five benchmark datasets. The table shows that our proposed GCBR method outperforms other baselines in terms of Micro-F1 and Macro-F1 on datasets of Pubmed, Reddit, Co-CS, and Co-Phy, and GCBR++ is the runner-up. Our methods have shown significant improvement in the imbalance ratio compared to the other baselines across all datasets. 
The advantages are more obvious when the class distribution is highly skewed. 
For instance, the datasets for Co-Phy and Co-CS exhibit the highest skewed class distribution among all the datasets. Our GCBR outperforms the best baseline in the metric of Macro-F1 by at least 2\% on Co-Phy and Co-CS. The imbalance ratios are improved by 31\% and 55\% with GCBR++ compared to the best baseline on Co-Phy and Co-CS. 
The superior performance of our GCBR and GCBR++ can be mainly attributed to our policy considering class balance. The class-balance-aware reward function and state space effectively enhance the performance of minority classes compared to other baselines, further improving overall performance, as shown in Figure. \ref{fig:fig1}. 
Although GCBR achieves the second-best Micro-F1 and Macro-F1 on Citeseer, it still demonstrates very competitive performance.
Furthermore, we also can find that GCBR++ achieves a higher imbalance ratio score but lower Macro-F1 and Micro-F1 compared to GCBR in the table. Because GCBR++ pays more attention to class balance by adding the punishment mechanism.

\subsubsection{Performance under Different Test Budgets}

\begin{figure}[!h]
\centering
\subfigure{
	\label{fig:RedditMicBudegt}
	\includegraphics[scale=0.32]{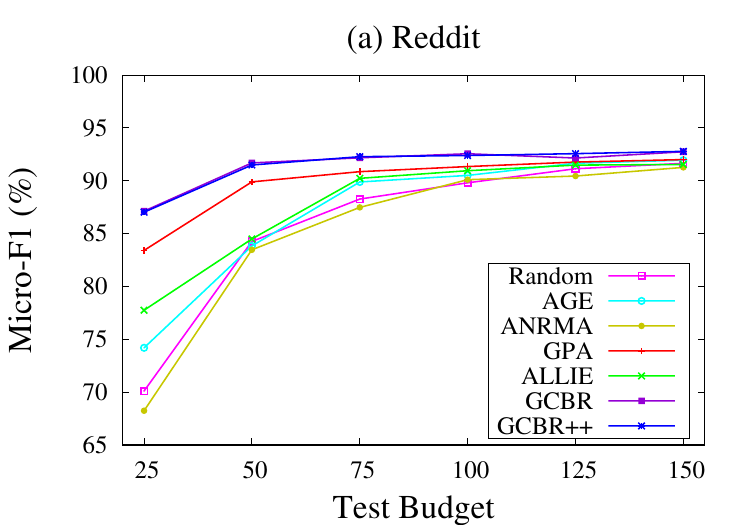}}
\hspace{0.01 in}
\subfigure{
	\label{fig:PhyMicBudegt}
	\includegraphics[scale=0.32]{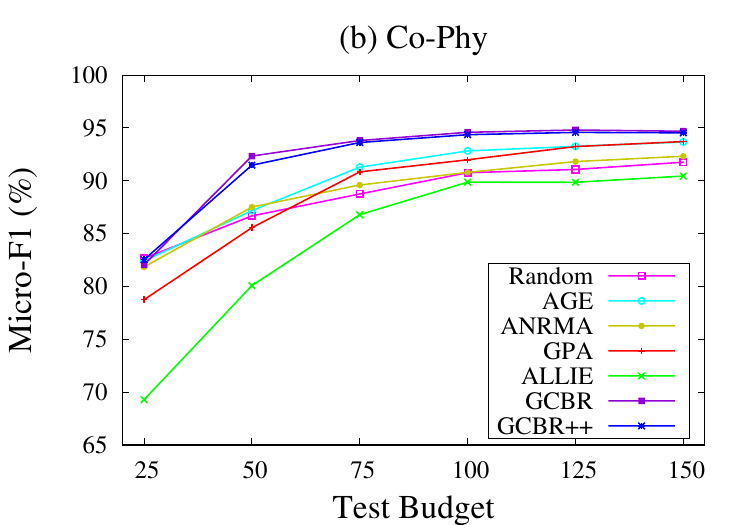}}

\subfigure{
	\label{fig:RedditMacBudegt}
	\includegraphics[scale=0.32]{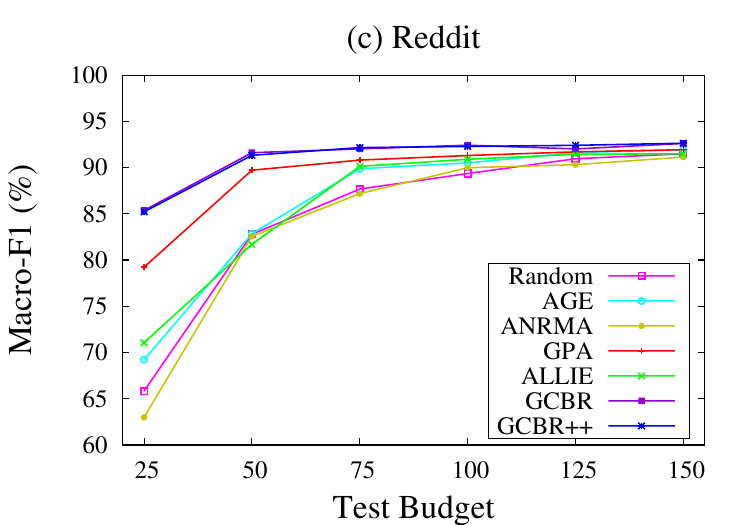}}
\hspace{0.01 in}
\subfigure{
	\label{fig:PhyMacBudegt}
	\includegraphics[scale=0.32]{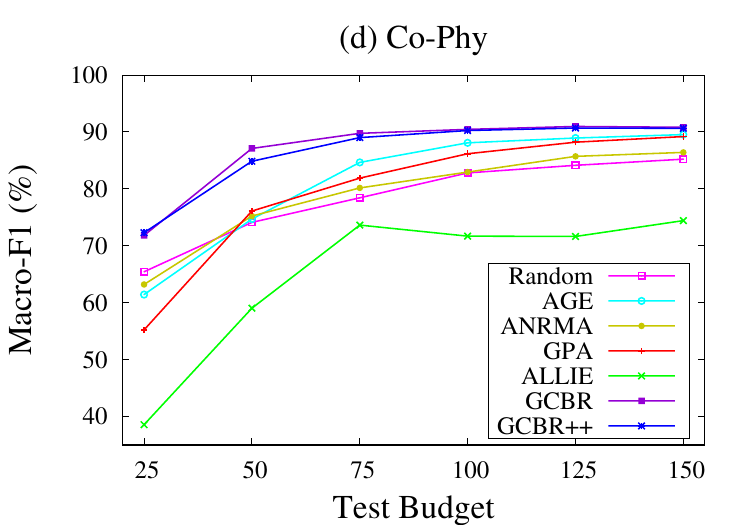}}

\subfigure{
	\label{fig:RedditRatioBudegt}
	\includegraphics[scale=0.32]{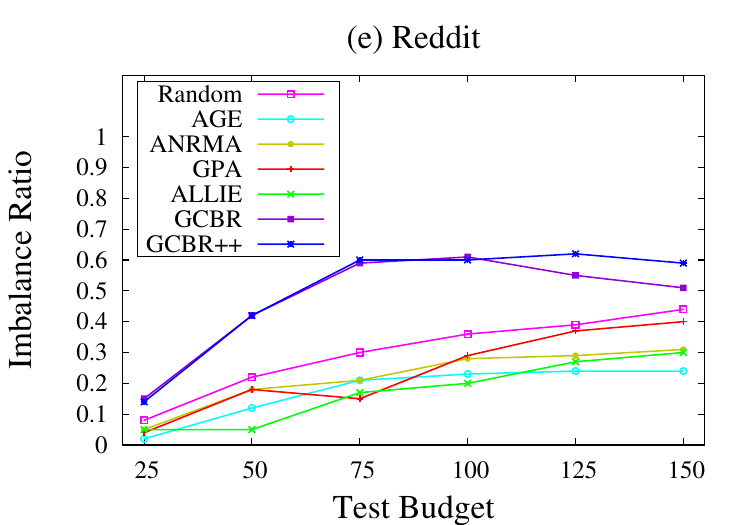}}
\hspace{0.01 in}
\subfigure{
	\includegraphics[scale=0.32]{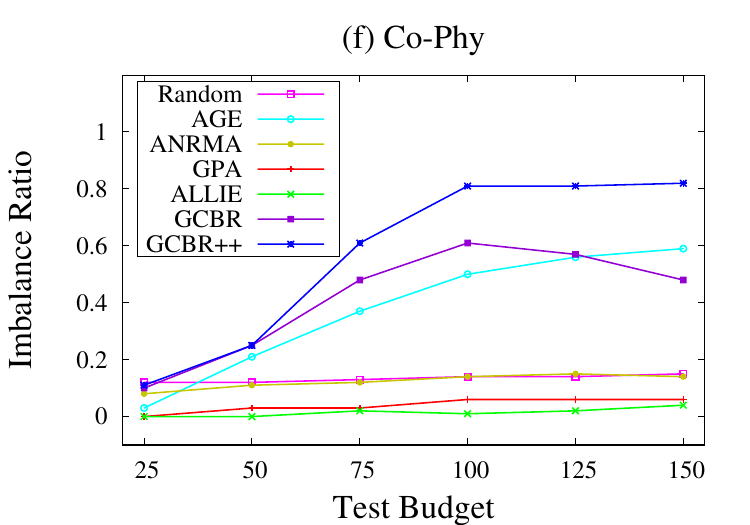}}

\caption{Node classification results (Macro-F1, Micro-F1) and Imb-ratio on Reddit and Co-Phy under different test budgets.}

\label{fig:performancebudgets}
\end{figure}

Here, we conduct experiments to compare all the methods under different test budgets. This study uses Reddit and Co-Phy as examples, representing datasets with relatively balanced and highly imbalanced class distributions, respectively. 
Figure \ref{fig:performancebudgets} (a)-(d) shows that our methods consistently outperform all baselines under all budgets in terms of Micro-F1 and Macro-F1 on both datasets. The benefits are more apparent when the test budget is small, and the class distribution is highly skewed. For instance, GCBR and GCBR++ outperform the best baseline by at least 6\% in terms of Macro-F1 on Reddit and Co-Phy when the number of labeled nodes is 25. When the test budget is 150, GCBR outperforms the best baseline by 1.28\% on Co-Phy but only by 0.68\% on Reddit in terms of Macro-F1.
This is because the number of nodes selected from minority classes by other baselines is very limited, which results in lower Micro-F1 and Macro-F1 scores. The smaller the test budget, the more skewed the class distribution, and the more pronounced this issue becomes.

Figure \ref{fig:performancebudgets} (e)-(f) shows that our methods consistently outperform all baselines under all budgets in terms of imbalance ratio on Reddit and Co-Phy. 
In specific, the imbalance ratio of GCBR and GCBR++ initially are positive to the test budgets, then converges for GCBR++ as the budget reaches around 75 and 100 on Reddit and Co-Phy, respectively, and decreases for GCBR slightly as the budget reaches around 100 on both datasets. 

GCBR and GCBR++ consistently outperform all baselines under different budgets in terms of Micro-F1, Macro-F1, and imbalance ratio. GCBR achieves better performance, while GCBR++ achieves competitive and more stable class balance under different budgets.

\subsubsection{Varied Scaling Factor}

\begin{figure*}[h]

\centering
\begin{minipage}[h]{0.3\textwidth}
\centering
\includegraphics[scale=0.33]{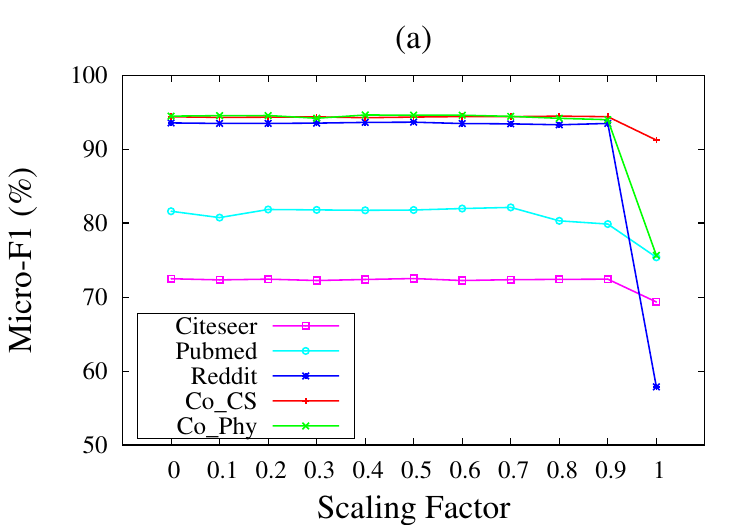}
\end{minipage}
\begin{minipage}[h]{0.3\textwidth}
\centering
\includegraphics[scale=0.33]{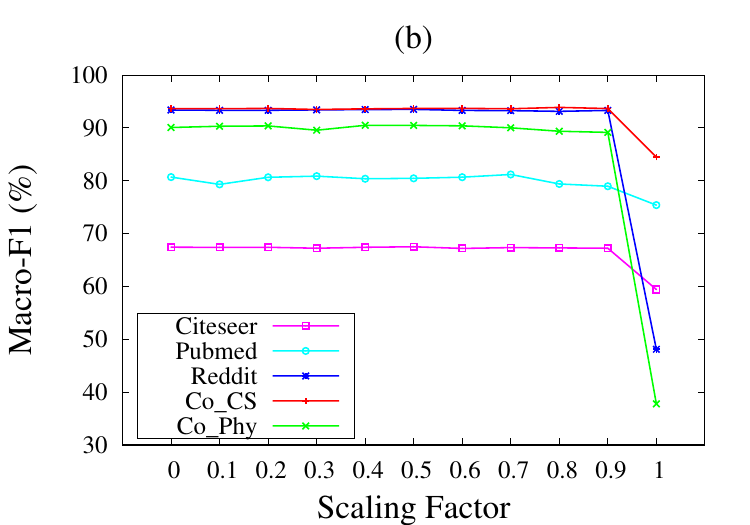}
\end{minipage}
\begin{minipage}[h]{0.3\textwidth}
\centering
\includegraphics[scale=0.33]{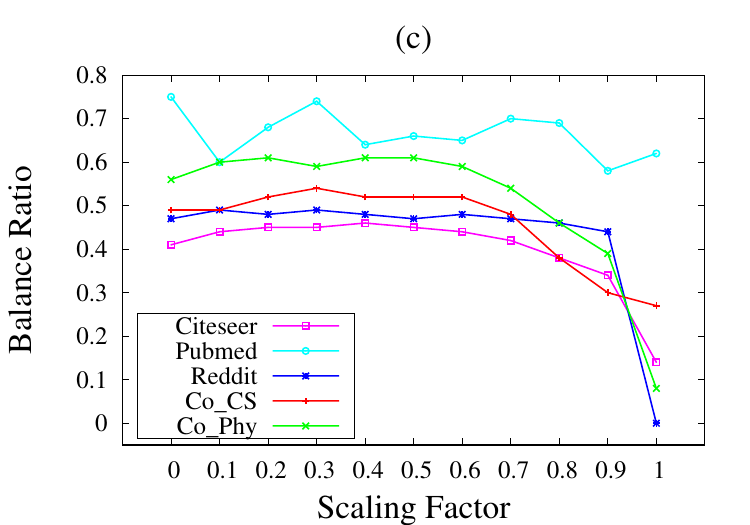}
\end{minipage}

\caption{The performance (Micro-F1, Macro-F1) and imbalance ratio of GCBR under different scaling factors $\alpha$.}
\label{fig:alphas}
\end{figure*}

In this section, we show the performance of GCBR under different scaling factor $\alpha$ on all five datasets. The results are illustrated in Figure \ref{fig:alphas}. We can find that Micro-F1, Macro-F1, and imbalance ratio fluctuate insignificantly when $\alpha$ is between 0 and 0.9, indicating that the performance of GCBR is not strictly sensitive to $\alpha$. Surprisingly, GCBR can achieve very good performance when the reward function only contains the class diversity part ($\alpha=0$). There is a significant performance drop when the reward function only includes the performance gain part ($\alpha=1$). Therefore, the class diversity we proposed in the reward function is crucial for our model's performance.

\subsubsection{Varied Penalty Score}

\begin{figure*}[h]

\centering
\begin{minipage}[h]{0.3\textwidth}
\centering
\includegraphics[scale=0.33]{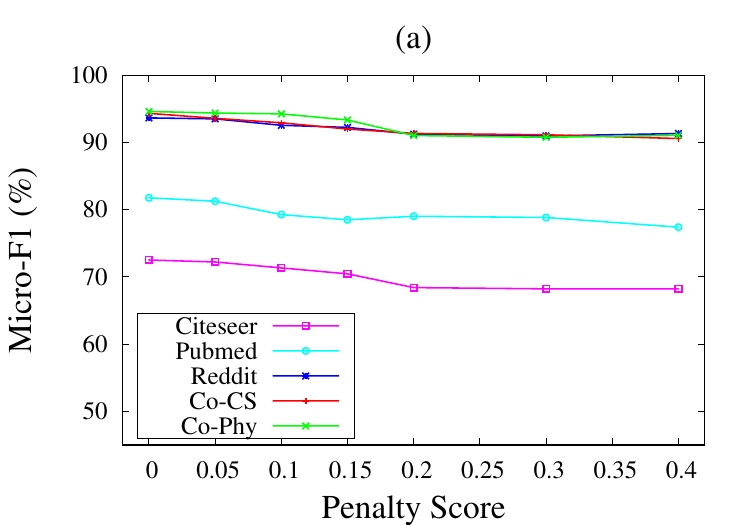}
\end{minipage}
\begin{minipage}[h]{0.3\textwidth}
\centering
\includegraphics[scale=0.33]{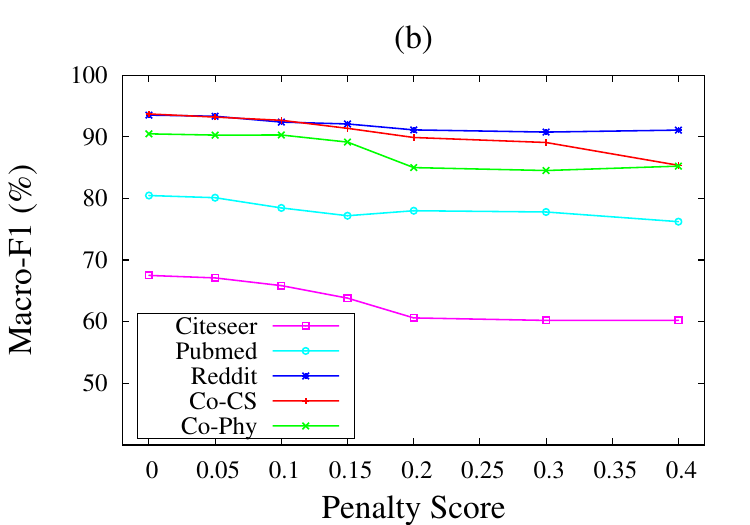}
\end{minipage}
\begin{minipage}[h]{0.3\textwidth}
\centering
\includegraphics[scale=0.33]{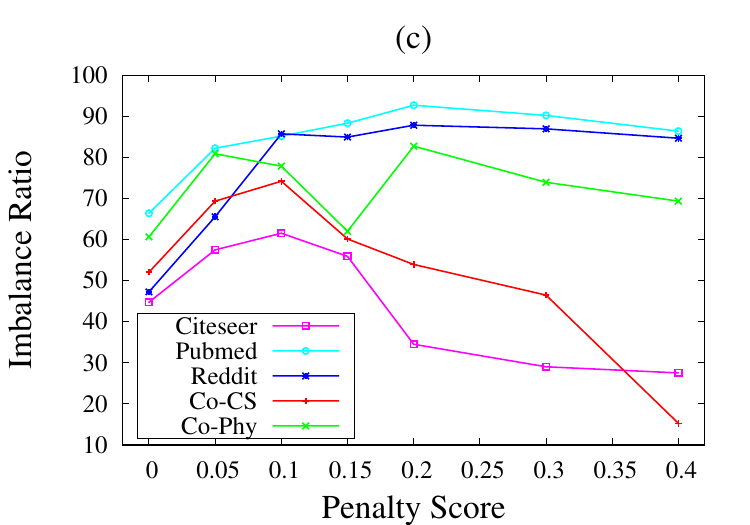}
\end{minipage}

\caption{The performance (Micro-F1, Macro-F1) and imbalance ratio of GCBR++ under different penalty scores $\eta$.}
\label{fig:penalty}
\end{figure*}

Figure \ref{fig:penalty} shows the performance and imbalance ratio of GCBR++ under different penalty score $\eta$. $\eta=0$ means GCBR++ is equal to GCBR. The performance of GCBR++ tends to degrade as $\eta$ increases for all datasets. The imbalance ratio of GCBR++ initially rises and then declines with the increase in the value of $\eta$. In GCBR++, we set $\eta=0.05$ as it corresponds to the minimal decrease in performance while exhibiting a significant improvement in class balance.

\subsubsection{Varied Training Budget}

\begin{figure*}[h]

\centering
\begin{minipage}[h]{0.3\textwidth}
\centering
\includegraphics[scale=0.33]{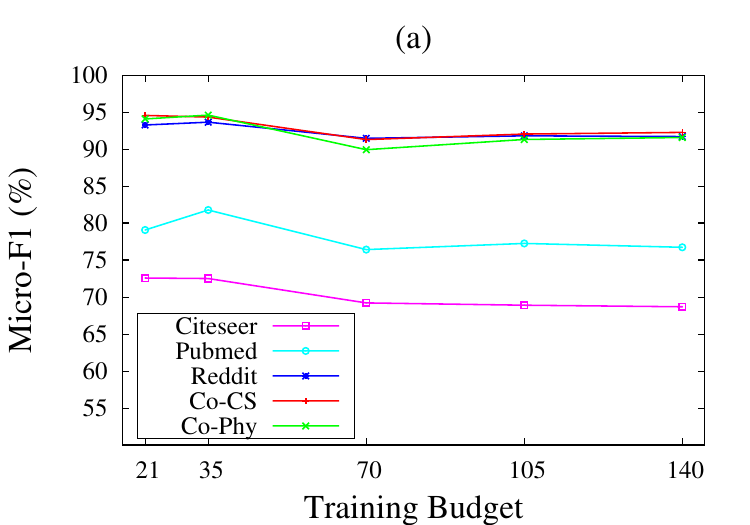}
\end{minipage}
\begin{minipage}[h]{0.3\textwidth}
\centering
\includegraphics[scale=0.33]{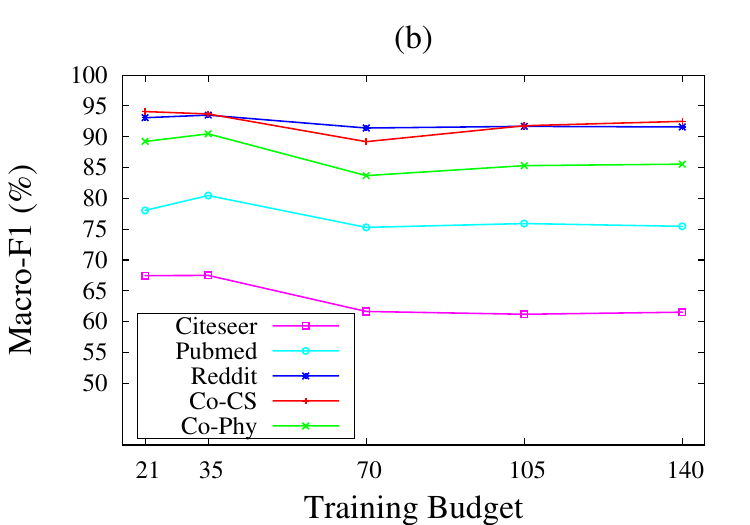}
\end{minipage}
\begin{minipage}[h]{0.3\textwidth}
\centering
\includegraphics[scale=0.33]{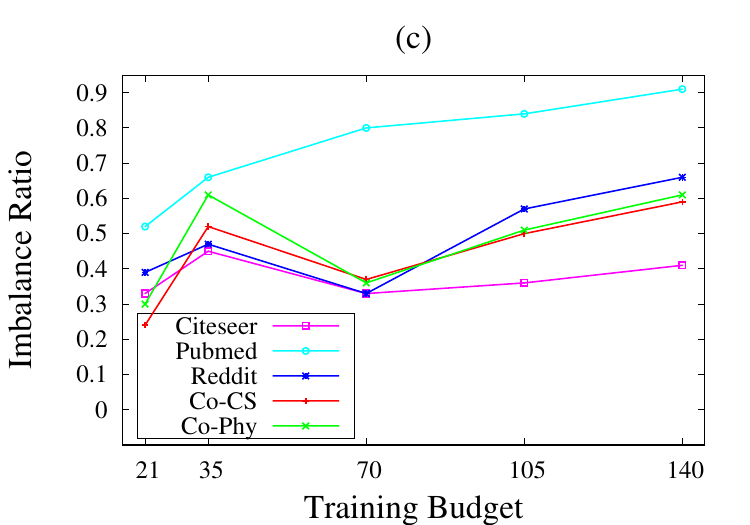}
\end{minipage}

\caption{The performance and imbalance ratio of GCBR trained with different training budgets.}
\label{fig:trainBudget}
\end{figure*}

Here, we explore the impact of varying training budgets on the performance of the learned policy. We train five policies with training budgets of \{21, 35, 70, 105, 140\} on Cora respectively, since Cora has 7 classes. Then, these policies are utilized to acquire nodes with a fixed test budget size (20 $\times$ \#classes) for the other five datasets. In Figure \ref{fig:trainBudget}, we observe that the training budget of 35 is sufficient to yield better performance and a higher imbalance ratio for all datasets, which is similar to GPA \cite{hu2020graph}.

\subsubsection{Ablation Study of State Features}

\begin{table*}
\caption{Contribution of each state feature. Each row corresponds to removing one feature. ``GCBR'' means using all the features.}
\setlength{\tabcolsep}{2pt}
    \centering
    \begin{tabular}{ccccccccccccccccc}
    \hline 
        \multirow{2}{*}{Method} & \multicolumn{3}{c}{Citeseer} & \multicolumn{3}{c }{Pubmed}& \multicolumn{3}{c }{Reddit}& \multicolumn{3}{c }{Co-CS}& \multicolumn{3}{c }{Co-Phy}\\
         & {\footnotesize Micro-F1} & {\footnotesize Macro-F1}  & {\footnotesize Imb-ratio} & {\footnotesize Micro-F1} & {\footnotesize Macro-F1}  & {\footnotesize Imb-ratio} & {\footnotesize Micro-F1} & {\footnotesize Macro-F1}  & {\footnotesize Imb-ratio} & {\footnotesize Micro-F1} & {\footnotesize Macro-F1}  & {\footnotesize Imb-ratio} & {\footnotesize Micro-F1} & {\footnotesize Macro-F1}  & {\footnotesize Imb-ratio}  \\
         \hline 
GCBR&	\textbf{72.50}&	\textbf{67.51}&	0.45&	81.76&	80.46&	0.66&	\textbf{93.64}&	\textbf{93.51}&	0.47&	\textbf{94.32}&	\textbf{93.69}&	0.52&	\textbf{94.59}&	\textbf{90.47}&	0.61\\
NoCSim&	72.39&	67.43&	0.46&	81.11&	79.91&	0.59&	93.23&	93.05&	0.50&	94.09&	93.67&	0.47&	94.05&	89.64&	0.59\\
NoUncer&	71.66&	66.43&	0.56&	77.16&	75.82&	\textbf{0.87}&	92.29&	92.19&	\textbf{0.76}&	93.37&	93.46&	\textbf{0.72}&	93.41&	89.19&	0.64\\
NoCentr&	72.22&	67.21&	0.44&	80.90&	79.50&	0.60&	93.29&	93.15&	0.48&	94.31&	93.62&	0.51&	94.32&	90.03&	0.59\\
NoSelec&	71.33&	66.50&	\textbf{0.69}&	\textbf{81.85}&	\textbf{80.92}&	0.70&	92.93&	92.85&	0.58&	93.16&	92.78&	0.65&	93.70&	89.26&	\textbf{0.77}\\
NoCDiv&	71.68&	66.48&	0.31&	78.77&	77.6&	0.33&	93.44&	93.27&	0.35&	93.62&	84.77&	0.05&	92.59&	87.74&	0.17\\

       \hline 
    \end{tabular}
    \label{tab:stateFeature}
\end{table*}

We also study the contribution of the state features introduced in Section \ref{states}. We remove each of them from the state space to observe how they affect the learned policy. 
We can draw the following observations from Table \ref{tab:stateFeature}. 
First, in most cases, removing any of the features can result in a performance drop w.r.t. Micro-F1 and Macro-F1 metrics, demonstrating the effectiveness of these features. 
Second, NoCDiv obtains the lowest imbalance ratios on all the datasets and also the lowest Macro-F1 scores on Co-CS and Co-Phy, which have
highly skewed class distribution. This indicates that the feature of class diversity is crucial for class balance and classification performance. 
For other methods, they all exhibit high class balance due to the presence of class diversity as state feature.
Third, we surprisingly find that, 
compared with NoCSim and NoCentr,
the removal of uncertainty (NoUncer) and selectivity (NoSelec) leads to a decrease in classification performance and a significant improvement in class balance. 
The reason could be that
the reward function consists of two parts: performance gain and class diversity score.
To maximize reward given a decreasing performance gain, 
the class diversity score will thus be increased.
Nevertheless, we have to clarify that overly-emphasizing class balance cannot guarantee the classification results.
Finally, GCBR achieves the best classification results and maintains a competitive class balance across all datasets, striking a good trade-off between performance and class balance.

\section{Conclusion}

In this paper, we propose GCBR, a novel reinforced class-balanced active learning for GNNs. GCBR learns an optimal policy to acquire class-balanced and informative nodes for annotation. The class-balance-aware state space and reward function are designed to trade-off between model performance and class balance. 
Besides, we further upgrade GCBR to GCBR++ by incorporating a punishment mechanism to obtain a more class-balanced labeled set. Experiments on multiple benchmark datasets demonstrate our methods can get a more class-balanced labeled set and achieve superior performance over SOTA baselines.

\bibliographystyle{ACM-Reference-Format}
\bibliography{sample-base}

\newpage

\appendix

\section{Pseudo-code}
In this section, we present the pseudo-code of our approach for policy training (Algorithm~\ref{alg:trainPolicy}).
\begin{algorithm}
\SetKwData{Left}{left}\SetKwData{This}{this}\SetKwData{Up}{up}
\SetKwFunction{Union}{Union}\SetKwFunction{FindCompress}{FindCompress}
\SetKwInOut{Input}{input}\SetKwInOut{Output}{output}
\Input{labeled training graph $G_s$, the training budget $B$, maximal training episode $E$, update frequency $F$}
\Output{query policy $\pi_{\theta}$}
Initialize parameters of the actor network $\pi_{\theta}$\;
Initialize parameters of the critic network $V_{\phi}$\;
\ForEach {episode $=1$ to $E$} { 
    $L^0= \emptyset$\;
    Initialize the classification GNN as $f(G_s,L^0)$\;
    \For {$t= 1$ to $B$} {
        Generate the state $S^t$ based on Equation \ref{equation:s1}-\ref{equation:s5}\;
        Sample an unlabeled node $v^t \sim \pi_{\theta}(S^t)$ from $V_{train}\textbackslash L^{t-1}$ and query for its label\;
        $L^t = L^{t-1} \cup \{v^t$\} \;
        Train GNN model $f(G_s,L^t)$ for one epoch \;
        Evaluate $f(G_s,L^t)$ on $V_{valid}$\;
        Get reward $r^{t}$ based on Equation \ref{equation:reward}\;
        Generate next state $S^{t+1}$\;
        Get the current and the next state values $V_{\phi}(S^{t})$ and $V_{\phi}(S^{t+1})$ using  critic\;
        Compute target $V^{*}(S^{t})$ and advantage $A(S^{t},v^{t})$ based on Equation \ref{equation:target} and \ref{equation:advantage}\;
        \If{$t$ mod $F = 0$}{
            Update critic $\pi_{\theta}$ by minimizing the loss based on Equation \ref{equation:critic}\;
            Update actor $V_{\phi}$ by minimizing the loss based on Equation \ref{equation:actor}\;
        }
    }
}
\caption{Train the policy using A2C}
\label{alg:trainPolicy}
\end{algorithm}

\vspace{20pt}

\end{document}